\documentclass[10pt,twocolumn,letterpaper]{article}

\usepackage{cvpr}              

\usepackage{graphicx}
\usepackage{amsmath}
\usepackage{amssymb}
\usepackage{booktabs}
\usepackage[accsupp]{axessibility}  

\def\lc{\left\lceil}   
\def\rc{\right\rceil}

\def\lf{\left\lfloor}   
\def\rf{\right\rfloor}

\usepackage[pagebackref,breaklinks,colorlinks]{hyperref}

\usepackage[capitalize]{cleveref}
\crefname{section}{Sec.}{Secs.}
\Crefname{section}{Section}{Sections}
\Crefname{table}{Table}{Tables}
\crefname{table}{Tab.}{Tabs.}

\def\cvprPaperID{10164} 
\def\confName{CVPR}
\def\confYear{2022}

\begin{document}

\title{Multimodal Colored Point Cloud to Image Alignment}

\author{Noam Rotstein \qquad Amit Bracha \qquad Ron Kimmel 
 \\
 Technion - Israel Institute of Technology
 \\
\tt\small\{snoamr, amit.bracha, ron\}@cs.technion.ac.il
}

\maketitle

\begin{abstract}
Reconstruction of geometric structures from images using supervised learning suffers from limited available amount of accurate data.
One type of such data is accurate real-world RGB-D images.
A major challenge in acquiring such ground truth data is the accurate alignment between RGB images and the point cloud measured by a depth scanner.
To overcome this difficulty, we consider a differential optimization method that aligns a colored point cloud with a given color image through iterative geometric and color matching.
In the proposed framework, the optimization minimizes the photometric difference between the colors of the point cloud and the corresponding colors of the image pixels.
Unlike other methods that try to reduce this photometric error, we analyze the computation of the gradient on the image plane and propose a different direct scheme.
We assume that the colors produced by the geometric scanner camera and the color camera sensor are different and therefore characterized by different chromatic acquisition properties.
Under these multimodal conditions, we find the transformation between the camera image and the point cloud colors.
We alternately optimize for aligning the position of the point cloud and matching the different color spaces.
The alignments produced by the proposed method are demonstrated on both synthetic data with quantitative evaluation and real scenes with qualitative results.
\end{abstract}

\section{Introduction}
\label{sec:intro}
In recent years, research in 3D shape reconstruction has made tremendous progress.
Much of the research in the field focused on shape reconstruction from IR or RGB images.
Multiple approaches have been proposed including amongst others, shape from stereo and monocular depth estimation.
More recently, the focus of the community shifted towards supervised deep learning methods \cite{zbontar2016stereo, kendall2017end, zhang2019ga, chang2018pyramid, luo2016efficient, pang2017cascade, eigen2014depth, eigen2015predicting, fu2018deep}.
Deep Learning relies heavily on large and accurate datasets.
Since such datasets are difficult to obtain, most works use synthetic datasets to train and validate their models.
However, these datasets are limited since they do not capture real-world properties such as distortions and noise.
In addition, the use of deep learning methods may require specific training data for each camera model, as the shape reconstruction algorithms are sensitive to model properties and artifacts.

One solution to acquire accurate ground-truth depth is to use precise, yet, often slow, depth measuring devices such as 3D laser scanners.
Using such a device to acquire exact depth values for a desired camera model requires registering the device pose relative to the camera one.
Placing the device at a fixed and calibrated position relative to the camera is often unsuitable for this task. 
Such a setup suffers from technical difficulties and requires constant maintenance \cite{geiger2013vision}. 
Accurate laser scanners usually require a long scanning time for a given scene.
Consequently, in order to create a large-scale dataset, multiple images should be acquired for each depth scan.
This means that a 3D Euclidean transformation must be found between each image coordinate system and the scanner coordinate system.
\begin{figure*}[htbp]
    \centering
    \includegraphics[width=1\textwidth]{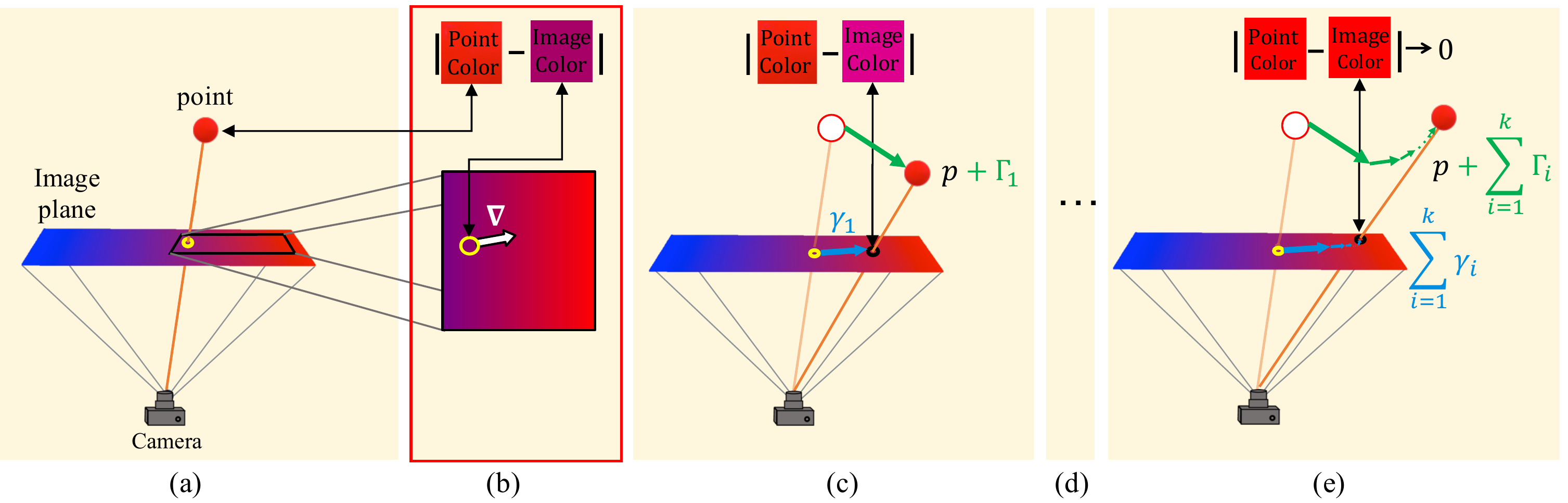}
    \caption{
   {
    The proposed framework operating, for simplicity, on a single point under a single modality setting.
    (a) Project the point onto the image plane.
    (b) Examine the photometric difference between the point color and the image color at the projected location. 
    Unlike projection, color on the image plane is not straightforward differentiable.
    Create a differential surface on the image plane.
    (c) Use the differentiation to compute the gradient of the translation parameters and perform a step in the direction of the gradient.
    (d) Move the point iteratively to optimize the photometric difference.
    (e) Stop when the photometric error between the point color and the image color at the projected point location is minimal. 
    In the case of many points in our point cloud, the update is also applied to the rotation parameters.
 }
    }
    \label{fig:onepoint}
\end{figure*}

The geometric information and color texture provided by some of these devices can be combined to produce a colored point cloud that can be used for registration.
However, such registration faces the challenge of multimodality. 
More specifically, with the comparison of two images taken with different devices - one with the scanner and one with the camera.
Different devices have different color properties, often referred to as color gamut.
When two devices capture the same scene, the captured color space of each device is different.
This defines the problem that our work attempts to solve.
How to accurately align a colored point cloud with a colored image under multimodal conditions.

The multimodal alignment task, which is critical to produce accurate and reliable geometric data, has not yet been directly explored.
While a rough alignment can be easily found, a precise alignment is required for a proper geometric dataset construction.
Therefore, in this paper, we focus on refining such an alignment.
Direct visual odometry methods can be modified to be used for our task.
Such methods have been mainly used in SLAM, Optical Flow and Color Mapping.
These approaches attempt to minimize the photometric error between corresponding pixels based on the estimated scene geometry.
However, unlike our task, these approaches align images acquired by the same device.

 In contrast to previous methods, the proposed pipeline uses a direct numerical sub-pixel scheme to approximate the gradients in the image plane.
We prove that the commonly used method for computing such gradients is equivalent to evaluating the gradients on a blurred image.
To overcome multimodality, we also propose a second-order polynomial color transformation between the point cloud colors and the image colors.
Such an approach has rarely been used for color transformations.
The proposed approach produces state-of-the-art results for multimodal colored point cloud to image alignment.
\section{Related Efforts} \label{RE}
Alignment and registration have been studied intensively in many different directions and setups.
One of the most fundamental tasks is 3D-to-3D alignment between two point clouds. 
The most common solution is the Iterative closest point algorithm (ICP) \cite{besl1992method, chen1992object}. 
Many improvements and variations of this algorithm have been proposed \cite{pomerleau2013comparing, bronstein2008rigid, mitra2004registration, jubran2021provably}. 
Other studies proposed to align the colored 3D point cloud using the additional RGB color information \cite{johnson1999registration, korn2014color, park2017colored} or hue values \cite{men2011color}. 

The task of 2D-to-2D image registration has also been widely explored and used in many applications.
The most popular approach is to find corresponding points in both images and then determine the transformation between them \cite{dawn2010remote}. 
The two leading methods for finding corresponding points are intensity-based methods and feature-based methods. 
Intensity-based alignment methods compare intensity patterns in images and image patches \cite{papademetris2004integrated, godin1994three}. 
Feature-based methods extract features in each image \cite{lowe2004distinctive, bay2006surf, arandjelovic2012three, rublee2011orb} and then match them.
Mapping between the image coordinates is then derived from the corresponding matches.


While 3D-to-3D and 2D-to-2D tasks have been thoroughly explored, we address a different challenge.
We are interested in fitting the colors of the 3D geometry to the 2D image.
Such a procedure \cite{kaminsky2009alignment}, aligns 3D point clouds to overhead images using edge costs and free space costs.
Visual-based localization (VBL), is a domain that attempts to approximate camera pose relative to known 3D models \cite{PIASCO201890}.
The most common method is to use image feature descriptors.
Features are extracted on a query image and compared to features coupled with 3D coordinates.
Then, registration is performed using the Perspective-n-Point (PnP) algorithm \cite{lu2000fast, lepetit2009epnp}.
In contrast to our task, which focuses on accurate registration, most VBL papers and benchmarks focus on efficient and fast matching between image features and features of large-scale geometric models \cite{sattler2011fast, liu2017efficient, feng20192d3d}. 

There are some approaches that can be associated with the proposed method for pose optimization.
Zhou and Koltun proposed to align multiple images to an uncolored point cloud \cite{zhou2014color}.
As opposed to the method we present, they require a few images, optimize a large number of parameters to find the colors of the points, and operate under a single modality setup.
Pulli et al. align two colored point clouds by minimizing color and range on two image planes \cite{pulli2005projective}, also under single modality assumptions.
These methods can be classified as direct visual odometry (DVO).
They optimize the geometry directly on the image intensities by minimizing the photometric reprojection error between images.
These techniques are used for camera localization \cite{kerl2013robust, steinbrucker2011real, engel2017direct}, simultaneous localization and mapping (SLAM) with RGB-D cameras \cite{kerl2013dense, audras2011real}, SLAM with stereo cameras \cite{engel2015large}, and SLAM with monocular cameras \cite{engel2014lsd}.
Some of the algorithms used for SLAM estimate the geometric model to perform registration.
Thus, although these algorithms attempt to register 2D images, their process can be related to our task.
A key difference between SLAM and the problem addressed is the multimodal configuration of the former.
In addition, we use a different direct numerical gradient approximation scheme that leads to a significant improvement in the alignment.
Other SLAM methods attempt to match between features in images in conjunction with their 3D coordinates \cite{campos2021orb}.
Unlike these methods, which optimize the geometric model while aligning images, we benefit from an accurate geometric model that can be used in our favor.

Some DVO methods attempt to perform affine lighting corrections \cite{engel2015large} or optimize gamma correction \cite{engel2017direct} while performing alignment.
However, these methods first convert the RGB values to grayscale.
We aim to operate in a multimodal environment and compare color values from different devices.
This comparison requires a color manipulation. 
Such manipulations have been widely explored, but have not yet been used for pose estimation.
The problem can be viewed as the gamut mapping problem, where the task is to find a transformation of color images from input to output devices.
Examples of such solutions are space-dependent gamut mappings \cite{mccann1999lessons, kimmel2005space}.
Sochen et al. show how different models of color perception, interpreted as geometries of the color space, lead to different enhanced processing schemes \cite{sochen1998general}.
Much of the work in this area focuses on the perceptual relationship between colors rather than their precise values.
For our concern, these solutions are not appropriate, since our problem requires the quantitative comparison of their values.
A classical approach to color manipulation of images is histogram equalization.
Caselles et al. try to overcome the fact that histogram modification sometimes does not produce good contrast by performing it locally on connected components of the image \cite{743856}.
These methods succeed in improving image contrast, but  were not designed for analytical comparison of color values.
Specifically, they do not consider color relations between corresponding pixels in different images.
Many papers attempt to perform color matching for value comparison when converting RGB signals to standard CIE tristimulus values.
Typical methods include three-dimensional lookup tables with interpolation \cite{hung1993colorimetric}, neural networks \cite{kang1992neural}, and polynomial regression models \cite{kang1992color}.
Lookup tables lack the differentiability necessary for our task.
In the method we propose, the corresponding colors to be matched are computed per iteration, so training a neural network each time is not a feasible solution.
On the contrary, polynomial regression models satisfy the necessary requirements for our goal.
Several experiments have investigated the influence of polynomial order on the success of color transformation \cite{hong2001study, yilmaz2004color}.
From these experiments, it can be concluded that the higher the order used, the better the results.
Practically, second order models proved to provide accurate transformations at low computational cost.

\section{Rigid Alignment and Color Matching}
In this section, we show how to directly align a colored point cloud to the perspective image of a given scene.
We assume a pinhole camera model with known intrinsic parameters.
Our model takes advantage of the fact that nearby pixels in natural images tend to have similar colors and that the color change is slow and gradual.
We leverage this property in an optimization scheme by moving the point cloud so that its colors and the colors of the image on the projected points locations match.
For simplicity, we first consider a case where the colors of the point cloud and the colors of the image were captured by the same device and share the same color gamut.
We denote the point cloud $\{x_j\}_{j=1}^n \in \mathbb{R}^{n \times 3}$ in the XYZ space, and its colors ${\{c_j\}}_{j=1}^n\in \mathbb{R}^{n \times 3}$, in the RGB space.
The 2D coordinates of a point projected onto the image plane are denoted by $p_j \in \mathbb{R}^2$. 
The colors of the image $I:\mathbb{R}^2\rightarrow\mathbb{R}^3$ at $p_j$ are given by $I(p_j)$.
Using the former notations, when a point is aligned, $c_j \approx I(p_j)$.
Note that unlike the image plane, the perspective projection of the point cloud onto the plane is a straightforward differentiable operation.
To obtain a fully differentiable procedure, we define a differentiable surface.
This surface is discussed and analyzed in Section \ref{sub_color_gradient}.
In the proposed procedure, the discrepancy depends on the 3D point coordinates.
For a one point scenario, we compute a 3D translation $\Gamma \in \mathbb{R}^3$ where the coordinates of the point in $\mathbb{R}^3$ are $x_j+\Gamma$.
Thus, the discrepancy is differentiable by $\Gamma$.
We use an iterative optimization procedure to find the translation $\Gamma$ that minimizes $|I(p_j+\gamma)-c_j|$ where $\gamma \in \mathbb{R}^2$ is the projected location of $x_j+\Gamma$ on the plane (see Fig \ref{fig:onepoint}).

Given a point cloud containing $n$ points, we repeat this operation for each point and move the point cloud as a whole to minimize the total color difference of the points.
To translate the points, we apply a 3D Euclidean transformation, namely translation and rotation.
The last transformation, denoted $T^{\theta}$, has six parameters (${\theta \in \mathbb{R}^6}$) that are iteratively updated during the optimization process.

\begin{figure}[htbp]
    \centering
    \includegraphics[width=0.48\textwidth]{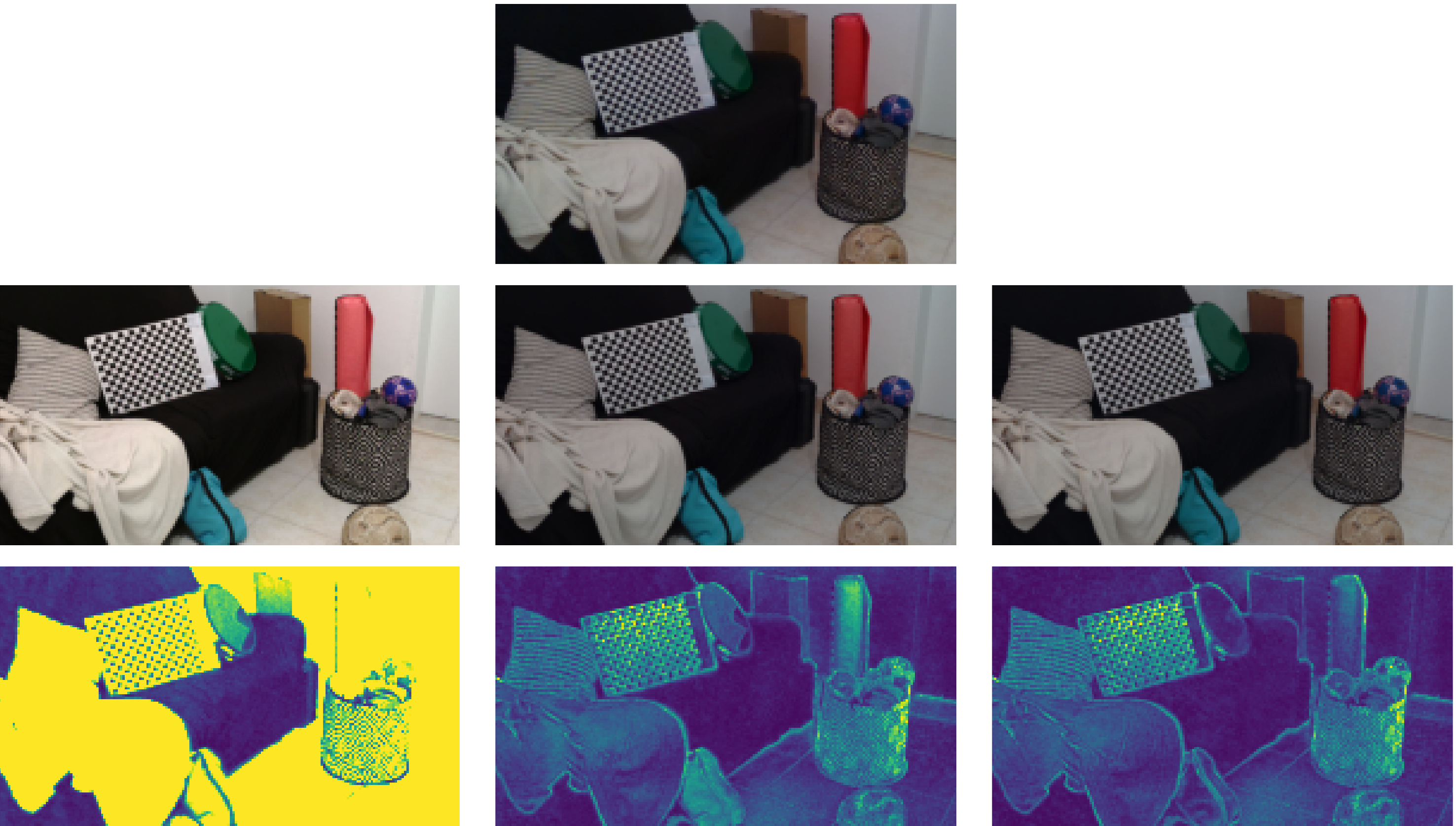}
    \caption{
    Top row - an RGB image captured by an Intel RealSense Depth Camera D435. Middle row - a rendered image from a point cloud scanned with a FARO 3D Focus Laser Scanner (left), the rendered image after a linear color transformation (middle), and the rendered image after a second-order polynomial transformation (right). The bottom row shows the absolute difference between the transformed modified image and the original image. Blue corresponds to small values, while yellow corresponds to large values. One can see that the color discrepancies of the linear transformation are reduced when using the second order polynomial transformation.
    }
    \label{fig:augment}
\end{figure}

In the former simple case described, we assumed that the point cloud and image had identical color gamut.
This assumption generally does not hold and images and point clouds captured with two distinct cameras sensing the same scene have different color values.
To get a meaningful and accurate alignment, we need to compensate for such color discrepancies between the two different types of sensors we are using.
Various DVO methods convert colors to grayscale and work with a single color channel.
In contrast, we propose to use the three-dimensional RGB space.
We attempt to compensate for color discrepancies without prior color manipulation or color calibration. 
At each iteration, we have a correspondence between the colors of the point cloud and the colors of the image.
That is, we have $3 \times n$ values to match.
One effective solution is to use this correspondence to find a linear transformation $D^{\mbox{linear}} \in \mathbb{R}^{3 \times 4}$ in the 3D color space.
This transformation fits between the two sets of colors.
However, in practice, a linear mapping of the colors cannot capture the complexity of color discrepancies \cite{hong2001study} (see Fig \ref{fig:augment}).
To this end, we propose the linear transformation $D \in \mathbb{R}^{3 \times 10}$ that minimizes the difference between second-order polynomials from one set of RGB colors to the other.
Although choosing a higher order polynomial provides higher accuracy, the marginal error between the second and higher order is small \cite{yilmaz2004color}.
Therefore, using second order provides the required accuracy while maintaining computational efficiency.

\subsection{Sub-Pixel Color}
A digital RGB image can be viewed as a discrete function
\begin{eqnarray}
J(a,b) \in[0,1]^3  & &a,b\in\mathbb{Z}.
\end{eqnarray}
Therefore, sub-pixel $u,v \in\mathbb{R}$ color values require interpolation.
We opt for the classical bilinear interpolation ($BL$),
\begin{eqnarray}
    I(u,v) &=& BL(J)(u,v).
\end{eqnarray}
\begin{figure*}[htbp]
    \centering
    \includegraphics[width=1\textwidth]{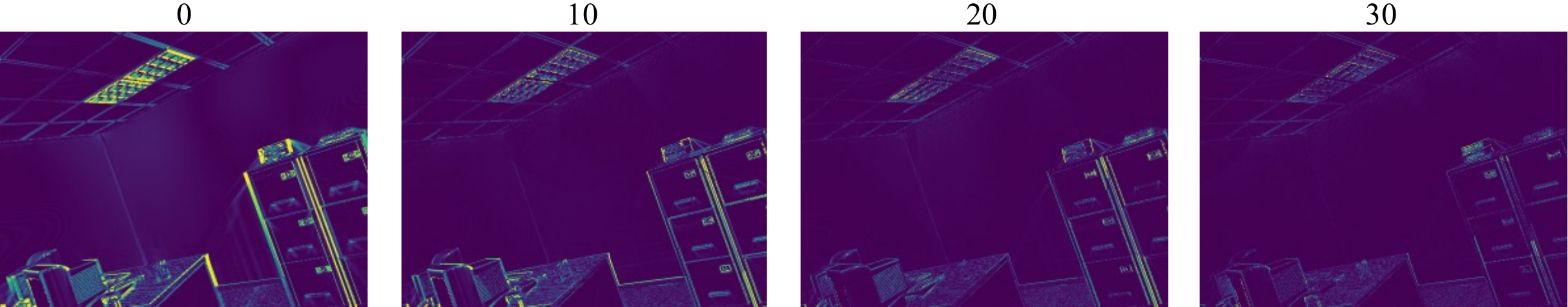}
    \caption{
    A visualization of the color difference during optimization iterations between an image and an image rendered from the point cloud. 
    Blue corresponds to small values, while yellow corresponds to large values. 
    One can see a significant difference in the initial non-aligned phase and the decrease of the difference as the optimization progresses.
    }
    \label{fig:convergence}
\end{figure*}
\subsection{Sub-Pixel Color Gradient} \label{sub_color_gradient}
Besides the gradient calculation in the image plane, our pipeline consists of straightforward differential steps. 
Given the discrete image $J$, we need to estimate the gradients between the pixel points.
Similar to the different definitions of pixel gradients \cite{kimmel2004level}, one can also use different definitions for sub-pixel gradients.
In this section we study two different definitions for such a calculation.
Strategy $A$, which is used in our method, and Strategy $B$, which is the the common calculation method in DVO implementations \cite{zhou2014color, kerl2013robust, kerl2013dense, audras2011real, engel2014lsd, engel2015large}.
\begin{enumerate}
    \item \textbf{Strategy $A$}- 
    The sub-pixel color values $I(u,v)$ can be viewed as a differential function of $u,v$.
    To obtain the gradient, this function is differentiated directly by $u$,
    \begin{eqnarray}
    I^{A}_{u}(u,v) &=& I_u(u,v) = BL(J)_u(u,v) \label{direct_diff_2d},
    \end{eqnarray}
    and similarly by $v$.
    \item \textbf{Strategy $B$}-
    First, the central finite difference approximation image of $J$ is computed,
    \begin{eqnarray}
    J_a(a,b) &=& \dfrac{J(a+1,b) - J(a-1,b)}{2}, 
    \label{discrete_2d_diff}
    \end{eqnarray}
    and similarly for $b$.
    Then, the sub-pixel accuracy is computed, by bilinear interpolation of the discrete gradient images,
    \begin{eqnarray}
    I^{B}_u(u,v) &=& BL (J_a) (u,v), 
    \label{interp_2d_diff}
    \end{eqnarray}
    and similarly for $v$. 
\end{enumerate}
Thus, the difference can be explained as follows, In strategy $A$ a continuous representation of the image is constructed and then differentiated, while in strategy $B$ a discrete gradient image is computed and then interpolated.
Let us analyze the two strategies using a 1-D example with linear interpolation of a discrete function $h(a)$, $a\in\mathbb{Z}$.
The sub-integer continuous interpolated values of the function at $x\in\mathbb{R}$ are,
\begin{eqnarray}
f(x) = (1-\delta) \cdot h(x_j) + \delta  \cdot h(x_{j+1}).
\end{eqnarray}
Where $f$ is the continuous function estimate, $x_j = \lf x \rf $, $x_{j+1} = \lc x \rc$ and $\delta= x - \lf x  \rf$.
In strategy $A$, as shown in Equation (\ref{direct_diff_2d}), the differentiation is done directly by $x$ and thus by $\delta$,
\begin{eqnarray}
f^{A}_x(x) = h(x_{j+1}) - h(x_j) \triangleq \Delta h_j. \label{our_diff}
\end{eqnarray}
Let us examine strategy $B$.
First, the gradients of the discrete function are computed according to Equation (\ref{discrete_2d_diff}), 
\begin{eqnarray}
h_a(a) = \dfrac{h(a+1) - h(a-1)}{2}.
\end{eqnarray}
Similar to Equation (\ref{interp_2d_diff}), linear interpolation is used to calculate the sub-pixel gradient:
\begin{eqnarray} \label{their_diff}
f^{B}_x(x) =& (1-\delta) \cdot h_a(x_j) + \delta \cdot h_a(x_{j+1}) = \\
=& \dfrac{(1-\delta) \cdot \Delta h_{j-1}+ \Delta h_j + \delta \cdot \Delta h_{j+1}} {2}. \nonumber 
\end{eqnarray}
The proof of the last transition and an extension to the 2D image domain is provided in the supplementary material.
Equations (\ref{our_diff}, \ref{their_diff}) can be used to relate the two strategies by the following convolution,
\begin{eqnarray}
f^{B}_x(x) = f_x^{A} * w (x).
\end{eqnarray}
Where $w$ is a rectangular window function,
\begin{eqnarray}
w(x) &=& 
\begin{cases}
0.5 ,&  -1\leq x \leq1\\
0, & \text{otherwise.}
\end{cases}
\end{eqnarray}
We conclude that the popular strategy $B$ for gradient computation is actually a smoothed version of the gradient computed by strategy $A$.
It is equivalent to applying the gradient proposed in strategy $A$ to a blurred image.
This, in turn, implies loss of high frequency information and gradients that are affected by values of distant pixels.
This simple, yet crucial, distinction between strategy $A$ and the commonly used strategy $B$ has a significant impact on the accuracy of the alignment.
The ablation study in Section \ref{abl_sec} demonstrates this important observation both empirically and quantitatively.

Our numerical approximation preferred strategy can be viewed from a different perspective.
The comparison between strategy $A$ and strategy $B$ relates to the difference between central finite difference approximations and forward or backward approximations.
See, for example, \cite{kimmel2004level}.
The reason for choosing central difference approximations lies in a numerical error evaluation derived from a truncated Taylor expansion. 
The approximation is relevant assuming that $\Delta x \ll 1$ , where $\Delta x$ is the sampling interval of the continuous image function.
However, if the information in the image involves high frequencies, this assumption may be misleading.
In such a case, the approximation error in deriving the numerical approximations of the derivatives is of an order of the change in the function they approximate.
We argue that high frequencies are crucial for accurate alignment and therefore tight numerical stencils are better suited for the task.
In most of our scenarios, we could assume that $\Delta x \gg 1$, in which the approximation would fail to properly capture the relevant numerical error.
For this particular case, 
there are better options than the central differentiation strategy $B$.
Indeed, a preferred option would be a tight numerical stencil that uses only $x_j$ and $x_{j+1}$ as in strategy $A$.

\subsection{Color Transformation}
The colors of the point cloud ${c}$ and the corresponding interpolated image colors ${I^{p}}$ of its points $p$ are obtained from different camera sensors.
Therefore, to compare them, we would like to find the proper color relation between them.
We assume that we can write each color of the image as a function of the colors of the corresponding point in the point cloud.
To approximate this unknown function, we apply a second-order polynomial kernel to the colors $\{I^{p}\}$,
{\small
\begin{eqnarray}
K(I^p) &=& K(R^{p}, G^p, B^p) \\  
&= & [1, R, G, B, RG, GB, RB, R^2, G^2, B^2] \in \mathbb{R}^{10 \times n}.\nonumber
\end{eqnarray}
}
In contrast to the framework of Hong et al. \cite{hong2001study} for camera colorimetric characterization,
we do not add the 3rd order term $RGB$ as an additional dimension.
The reason is that the experimental results have shown no significant advantage when this dimension is added.
The point cloud alignment improves in each iteration.
Therefore, the correspondence between the color values of the point cloud and the color values of the image improves as well.
To exploit this, we find the color transformation repeatedly for each iteration $i$.
Then, the transformation is applied to compute the transformed colors derived from the image
\begin{eqnarray}
   I^{D_i} &=& D_i\, K(I^{p}). 
\end{eqnarray}
\subsubsection{Color Transformation Optimization}
To avoid outliers affecting the color transformation, a scheme for affine illumination correction \cite{engel2015large} is used.
An inlier point is defined as a point that holds,
\begin{eqnarray}
   \left \|I^{D_i}_j -  c_j\right \| &<& \beta_{max}.
   \label{color_transform_tresh}
\end{eqnarray}
The series of coefficients $D_i \in \mathbb{R}^{3 \times 10}$ of the polynomial terms that minimizes the sum of the color differences of the inliers is computed by the least squares method,
\begin{eqnarray}
   D_i& =& \arg_D \min \left \|I - c\right \|^2.
   \label{find_color_transform}
\end{eqnarray}
The combined optimization problem for finding the inliers and computing $D_i$ is solved alternatively and iteratively.
In the supplementary, we show how to handle color values exceeding $[0,1]$ while preserving differentiability.

\subsection{Proposed Scheme}
The complete procedure for each iteration ${i}$ can be described by the following steps.
\begin{enumerate}
    \item Transform the coordinates $x$ with the current transformation $T^{\theta_{i}}$.
    \item Apply a Z-buffer $E$ to mask out occluded points.
    \item Project the 3D coordinates onto the image plane using perspective projection $\mbox{Proj}$.
    \item Calculate the interpolated image colors of the projected points using $I$.
    \item Lift interpolated image colors with a second order polynomial kernel $K$.
    \item Find the color transformation $D_i$ between kernel applied colors and point cloud colors and transform colors.
\end{enumerate}
The computed color values depend on ${\theta_{i}}$,
\begin{eqnarray}
    I^{*_i} &=& D_i K \left (I\left(\mbox{Proj}\left(E\left(T^{\theta_{i}}(x)\right)\right)\right)\right) \in \mathbb{R}^{3 \times n} \label{pipeline_eq}.
 \end{eqnarray}
 
\subsection{Alignment Optimization}
We can define the photometric error for each point and for each color channel $l$,
\begin{eqnarray}
r_{jl} = I^{*_i}_{jl} - c_{jl}.
\end{eqnarray}
Using the above definitions, the loss function can be calculated.
A weighted nonlinear least squares approach is chosen,
\begin{eqnarray}
L(\theta_i) &=& \left \| I^{*_i} - c \right\|^2_{W}\, =\, \sum_{j=1}^{n} \sum_{l=1}^{3} w_{jl}r_{jl}^2. 
\end{eqnarray}
The weights $w_{jl}$ are computed according to the proposed non-Gaussian error models \cite{kerl2013robust}, which assume t-distributed errors with $\nu=5$ degrees of freedom,
\begin{eqnarray}
w_{jl} = \dfrac{\nu+1}{\nu+\frac{r_{jl}^2}{\sigma^2}}.
\end{eqnarray}
The value of $\sigma$ is computed iteratively,
\begin{eqnarray}
\sigma = \dfrac{1}{3n}\sum_{j=1}^{n} \sum_{l=1}^{3} r_{jl}^2 \dfrac{\nu+1}{\nu+\frac{r_{jl}^2}{\sigma^2}}.
\end{eqnarray}
We implemented and optimized the proposed algorithm with Pytorch and Pytorch Autograd and used the Adam algorithm for optimization.
Since the angle and translation parameters are of different units and orders of magnitude, we initialize their learning rate accordingly. 
For a megapixel image, the method converges in a few hundred iterations (see Fig \ref{fig:convergence}), which takes about $30$ seconds on a single GeForce GTX 2080ti GPU.
Since runtime or memory constraints are not critical to dataset construction, we do not focus on an efficient implementation.

\section{Experiment setups}
In this section we show how the proposed method performs in two scenarios.
The first, involves a synthetic database. 
By controlling the image generation process, we can accurately quantify the accuracy of the method.
Next, we apply the proposed method to real images.
We demonstrate success in aligning a FARO laser scanner point cloud with an image captured by an RGB camera.
The alignment is evaluated qualitatively by projecting the intensity edges of the rendered point cloud onto the edges of the given image.
\begin{figure*}[htbp]
    \centering
    \includegraphics[width=0.9\textwidth]{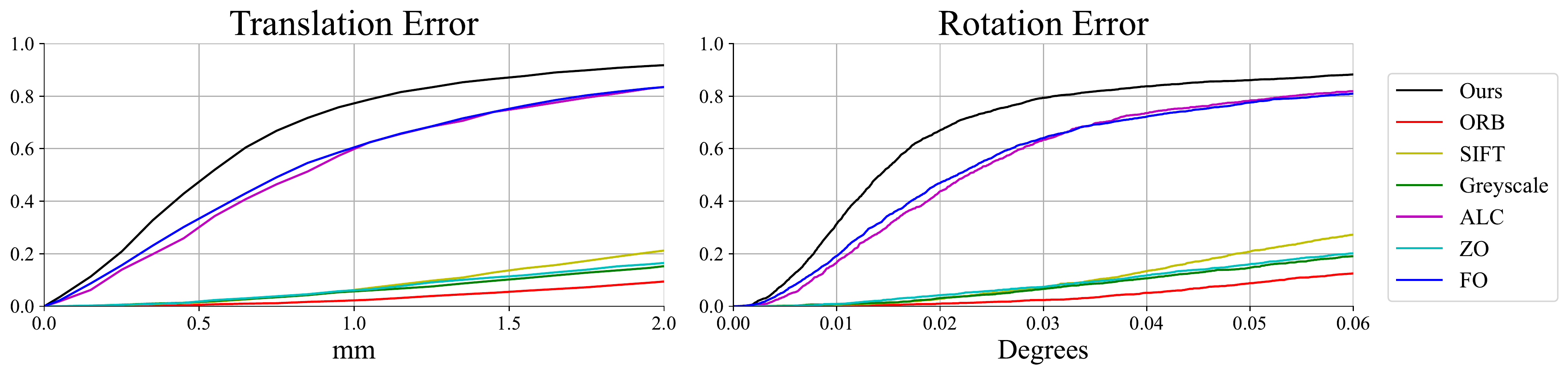}
    \caption{
    Cumulative normalized histograms of translation and rotation errors for synthetic data experiment. 
    The proposed method with second-order polynomial color alignment outperforms the rest of the methods.
    }
    \label{fig:syntheticResults}
\end{figure*}

\subsection{Synthetic Data} 
\label{synth_exp}
To demonstrate the proposed method, a photorealistic dataset is needed. 
We use the synthetic dataset ICL-NUIM \cite{handa2014benchmark}, which contains significant synthetic depth and RGB noise models representing a realistic environment.
From the dataset, 2000 images and their corresponding consecutive images are randomly sampled.
A corresponding point cloud of the images is created using their corresponding depth values.
These point clouds are misaligned with the successive images.
To simulate multimodality, we apply a series of effects to each of the RGB images. 
Many effects can be applied, such as effects that take into account spatial color configuration \cite{kimmel2005space} or perform retinex modifications \cite{kimmel2003variational, elad2003reduced, land1971lightness}.
To simplify the experiment, a set of elementary effects was chosen
Their numerical values with an example are discussed in the supplementary material,
\begin{enumerate}
    \item Apply random color transformation of brightness, contrast, saturation, and hue.
    \item Apply gamma correction with a random gamma value.
    \item Simulate different point spread functions and sensor properties by applying a Gaussian blur to the image.
\end{enumerate}
Point cloud to image alignment has no direct comparison.
Therefore, we study the use of modifications of different works,
\begin{enumerate}
    \item \textbf{ORB} - Feature-based camera localization with Oriented FAST and Rotated BRIEF (ORB) \cite{rublee2011orb} features is one of the most popular approaches for SLAM \cite{campos2021orb}.
    This approach is based on frame-to-frame registration in conjunction with coupled depth values.
    To modify it for our task, an image is rendered from the point cloud.
    ORB features are found on the original image and the rendered image.
    The point cloud $3D$ coordinates are used to determine the $3D$ coordinates of the features found on the rendered image.
    After matching the feature descriptors between the images, the Ransac algorithm is applied together with the PnP algorithm to compute the Euclidean transformation.
    \item \textbf{SIFT} - A common methodology in VBL implementations is to use Root Sift features \cite{lowe2004distinctive, arandjelovic2012three}. After the features are found and matched, the alignment is performed  as in using ORB features.
    \item \textbf{Greyscale} \label{gs_comp} -DVO methods commonly use grayscale images \cite{kerl2013robust, steinbrucker2011real} instead of RGB images. In this setup, we convert the colors to grayscale and then perform the alignment.
    \item \textbf{Affine Lighting Correction (ALC)}\label{alc_comp} - Engel et al. \cite{engel2015large} propose to first convert the colors to grayscale. Then, an affine lighting Correction is performed by alternately optimizing two parameters. These parameters form an affine intensity transform that corrects the grayscale values.
    \item \textbf{Zero Order (ZO)}\label{zo_comp} - Our pipeline used with no color transformation with direct comparison of the RGB color values.
    \item \textbf{First Order (FO)}\label{fo_comp} - Our pipeline used with first-order color transformation instead of second order.
\end{enumerate}
It is important to note that methods \ref{gs_comp}, \ref{alc_comp}, \ref{zo_comp}, \ref{fo_comp} have been modified to use our scheme for computing gradients.
This scheme is tested separately in Section \ref{abl_sec}.

The error is captured by two measures, the translation error and the rotation error.
The translation error is the Euclidean distance between the original translation and the derived translation.
To calculate the rotation error, the combined rotation axis is found using the rotation error about each axis.
The error is calculated by computing the rotation about each axis.
We present the cumulative normalized histogram of these errors under the different configurations in Fig. \ref{fig:syntheticResults}. 
Our approach clearly outperforms the other configurations.
Linear color transformations (FO and ALC), while beneficial, are inferior to the second-order color transformation.
Feature-based approaches (ORB and SIFT) and approaches that do not perform color transformation (ZO and Greyscale) achieve worse results compared to the alternatives.
We speculate that accurate geometry leads DVO methods with color modifications to significantly surpass feature-based approaches.
We denote that with the ICL-NUIM dataset and our method, we achieve a median sub-millimeter translation error of $0.58$mm and a median rotation error of $0.014^{\circ}$.

\subsection{Real Data}
To show that the proposed method works on real data, we used a FARO 3D Focus Laser Scanner and an Intel RealSense Depth Camera D435. 
We only use the RGB image from this camera and not its depth sensing capabilities. 
After each scan is completed, we place a camera near the scanner position. 
As mentioned earlier, the method relies on a coarse estimate of the camera pose.
We place a checkerboard in the scene and use it to roughly estimate the camera pose relative to the scanner. 
This will be the initial transformation we use before applying the proposed method. 

\begin{figure}[htbp]
    \centering
    \includegraphics[width=0.5\textwidth]{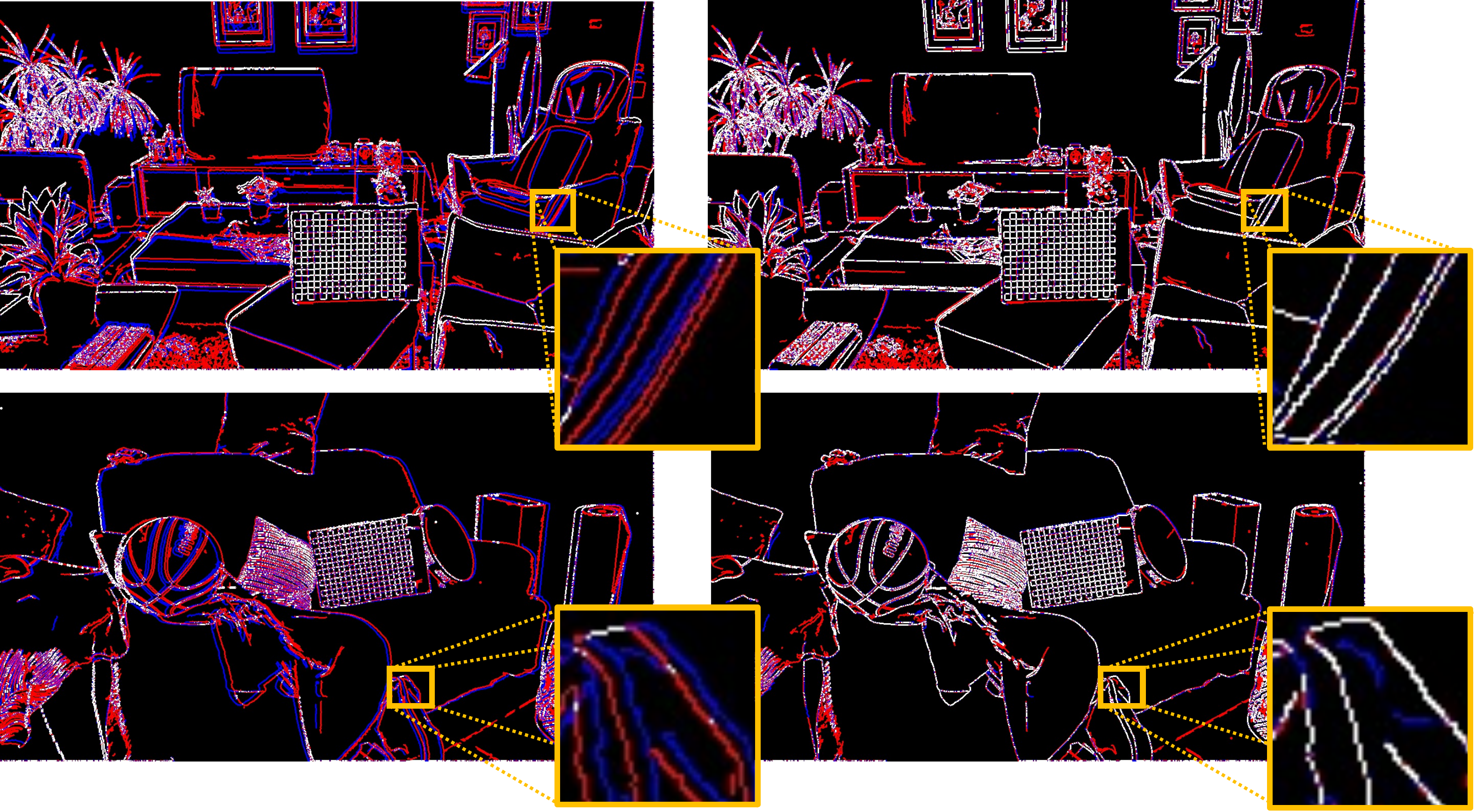}
    \caption{A comparison between edges from camera images and edges extracted from rendered point cloud images. Left is at initialization, right - proposed refinement. The colors denote the origin of the edges: red- camera image, blue- rendered image from point cloud, white- overlap. In the initial rough alignment, the corresponding edges do not overlap.
    }
    \label{fig:realResults}
\end{figure}

For simplicity, we use a checkerboard pattern.
Such an initialization can also be found using one of the methods described in Section \ref{RE}. For example, by applying a coarse-to-fine scheme \cite{steinbrucker2011real}.
Unlike the scenario with the synthetic data, we do not have the ground truth transformation. 
Therefore, we estimate the success of the proposed method  by visually comparing the RGB image and the rendered image from the point cloud after applying the computed transformation. 
Since different cameras are used, a simple color difference is not a good visual measure. 
Image edges are less affected by the camera characteristics and reflect whether the images are aligned correctly. 
Therefore, we find the edges of each image using the Canny-Haralick edge detector \cite{haralick1984digital, canny1986computational, kimmel2003regularized} and compare the edge images.
Fig. \ref{fig:realResults}, shows the edges extracted from both rendered and camera images, see the supplementary material for more examples. 
Due to the imperfect extraction of edges, not all edges in each image are detected. 
We can see that the edges in both images are aligned. 
As we can observe, this is in contrast to the edge comparison of the first misalignment of the edges. 
Although we cannot quantify the exact error in the real data scenario, the edge representation shows how accurate the proposed method is.

\subsection{Ablation Study} \label{abl_sec}
\begin{figure}[!htbp]
    \centering
    \includegraphics[width=0.48\textwidth]{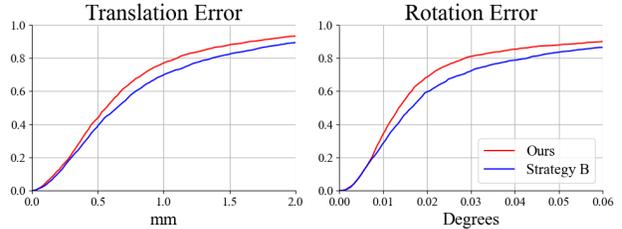}
    \caption{
    Cumulative normalized histograms of translation and rotation errors for the ablation study experiment. The proposed method for computing sub-pixel gradients on the image plane outperforms the common approach.
    }
    \label{fig:ablation}
\end{figure}

We demonstrate the impact of our sub-pixel gradient computation using the synthetic experiment setup.
We perform a test of our method with a single difference.
The calculation of the sub-gradient is changed.
The alternative configuration uses the usual gradient calculation in DVO implementations as described in Section \ref{sub_color_gradient}.
From the experimental results (Fig. \ref{fig:ablation}), such a configuration leads to inferior alignment compared to our method.
The median translation error increases by $13.2\%$ and the median rotation error increases by $18.4\%$.

\section{Limitations}
As shown, our sub-pixel gradient computation benefits from high-frequency detail and improves alignment accuracy in the ICL-NUIM dataset.
Although the dataset contains a simulation of real world noise, we speculate that extremely noisy scenarios may benefit from blurring and loss of such detail.
However, we believe that such blurring should be intentional.
The presented method may suffer from additional limitations common to DVO methods.
For example, if the camera is distant from the scanner, the point cloud may contain missing parts that do not appear on the image, or vice versa.
This could potentially degrade the alignment results.

\section{Conclusions}
We introduced an iterative differential method that aligns a colored point cloud to an image in a multimodal environment using geometric and second-order polynomial color matching and gradient-based optimization.
The proposed framework introduces an algorithmic pipeline that uses the entire point cloud and image information to minimize the discrepancy between the point cloud colors and projected image colors.
We analyze the computation of the gradient on the image plane and show an efficient and direct form of computing it.
We explain and numerically support the advantages of using second- order polynomials for color transformation between different camera devices.
We believe that the proposed concepts could facilitate and improve the creation of real 3D datasets in the future and could be applied to any camera model.
\newline
\newline
\textbf{Acknowledgements}-
This work addresses a challenge we were presented with during an internship at Intel RealSense.
We would like to thank Hila Eliyahu Grosman and Aviad Zabatani for their assistance.
The research was supported by the D. Dan and Betty Kahn Michigan-Israel Partnership for Research and Education, run by the Technion Autonomous Systems and Robotics Program.



{\small
\bibliographystyle{ieee_fullname}
\bibliography{egbib}
}

\end{document}


\title{Multimodal Colored Point Cloud to Image Alignment- \\Supplementary Material}

\maketitle

\section{Clipping and Differentiability}
The color transformation in section (3.3) can turn color values into values that exceed $[0,1]$.
This could introduce a bias to the comparison with $c$, which is within bounds.
We use a simple clip operation for each color value $V$,
\begin{align*}
V = \min(\max(V ,0 ),1 ).
\end{align*}
However, the use of this clipping method results in the clipped values no longer depending on $\theta_i$ and therefore, undesirably not being included in the optimization process.
To fix this, the original gradient before clipping is used for the optimization process.
In this way, we keep the differentiability and gradients while truncating the problematic color values.
The clipping operation is then performed on the transformed colors.

\section{Transition Proof}\label{transition_proof}
A proof of the last transition of Equation (9),
\begin{eqnarray*}
f^{B}_x(x)&=&(1-\delta) \cdot h_a(x_j) + \delta \cdot h_a(x_{j+1})  \cr
&=&
\overbrace{\dfrac{\Delta h_j}{2}-\dfrac{\Delta h_j}{2}}^0+ (1-\delta) \cdot h_a(x_j) + \delta \cdot h_a(x_j+1) \cr
& =& \dfrac{\Delta h_j}{2} - \dfrac{h(x_{j+1})-h(x_{j})}{2} + (1-\delta) \cdot \dfrac{h(x_{j+1})-h(x_{j-1})}{2} + \delta \cdot \dfrac{h(x_{j+2})-h(x_{j})}{2} \cr
&=& \dfrac{\Delta h_j}{2} + (1-\delta) \cdot \dfrac{h(x_{j})-h(x_{j-1})}{2} + \delta \cdot \dfrac{h(x_{j+2})-h(x_{j+1})}{2} \cr
& =&\dfrac{(1-\delta) \cdot \Delta h_{j-1}+ \Delta h_j + \delta \cdot \Delta h_{j+1}} {2} \cr
&=& f_x^{A} * w (x).
\end{eqnarray*}
The last transition can be easily derived from the definition of convolution.
\section{Extension to 2D and Bilinear Interpolation}
{
Let us analyze strategies $A$ and $B$ using the 2-D image and bilinear interpolation.
First, we study the partial derivative by $u$ with strategy $A$, we denote $u = u_j+\delta_u$,  $v = v_k+\delta_v$,

\begin{align*}
I^{A}_{u}(u,v) &= BL(J)_u(u,v) \\ 
& =\dfrac{d}{du}
\left(
\begin{bmatrix}
1-\delta_u & \delta_u 
\end{bmatrix}
\begin{bmatrix}
J(u_j,v_k) & J(u_{j},v_{k+1}) \\
J(u_{j+1},v_{k}) & J(u_{j+1},v_{k+1}) 
\end{bmatrix}
\begin{bmatrix}
1-\delta_v \\ \delta_v 
\end{bmatrix}
\right) \\
& =\begin{bmatrix}
\dfrac{d}{du}(1-\delta_u) & \dfrac{d}{du}(\delta_u)
\end{bmatrix}
\begin{bmatrix}
J(u_j,v_k) & J(u_{j},v_{k+1}) \\
J(u_{j+1},v_{k}) & J(u_{j+1},v_{k+1}) 
\end{bmatrix}
\begin{bmatrix}
1-\delta_v \\ \delta_v 
\end{bmatrix}\\
& =\begin{bmatrix}
-1 & 1 
\end{bmatrix}
\begin{bmatrix}
J(u_j,v_k) & J(u_{j},v_{k+1}) \\
J(u_{j+1},v_{k}) & J(u_{j+1},v_{k+1}) 
\end{bmatrix}
\begin{bmatrix}
1-\delta_v \\ \delta_v 
\end{bmatrix} \\
& =\begin{bmatrix}
J(u_{j+1},v_{k}) - J(u_j,v_k) &
J(u_{j+1},v_{k+1}) - J(u_{j},v_{k+1})
\end{bmatrix}
\begin{bmatrix}
1-\delta_v \\ \delta_v
\end{bmatrix}
\\
& \triangleq\begin{bmatrix}
\Delta_a J_{j,k} &
\Delta_a J_{j,k+1}
\end{bmatrix}
\begin{bmatrix}
1-\delta_v \\ \delta_v
\end{bmatrix}
\\
& =(1-\delta_v) \cdot \Delta_a J_{j,k} + \delta_v \cdot \Delta_a J_{j,k+1}
\end{align*}
Using strategy $B$,
\begin{flalign*}
& & I^{B}_u(u,v) &=BL (J_a) (u,v) \\
& & &=\begin{bmatrix}
1-\delta_u & \delta_u 
\end{bmatrix}
\begin{bmatrix}
J_a(u_j,v_k) & J_a(u_{j},v_{k+1}) \\
J_a(u_{j+1},v_{k}) & J_a(u_{j+1},v_{k+1}) 
\end{bmatrix}
\begin{bmatrix}
1-\delta_v \\ \delta_v 
\end{bmatrix}
\\ 
& & &=\begin{bmatrix}
\overbrace{(1-\delta_u) \cdot J_a(u_j,v_k)  + \delta_u \cdot J_a(u_{j+1},v_{k})}^{E_1} &
\overbrace{(1-\delta_u) \cdot J_a(u_{j},v_{k+1})  + \delta_u \cdot J_a(u_{j+1},v_{k+1})}^{E_2}\\
\end{bmatrix}
\begin{bmatrix}
1-\delta_v \\ \delta_v.
\end{bmatrix}
\end{flalign*}
}
Without the loss of generality, we analyze $E_1$,
Since $E_1$ is a 1-D function of $\delta_u$, we can use the proof \ref{transition_proof},
\begin{align*}
E_1 &=(1-\delta_u) \cdot J_a(u_j,v_k)  + \delta_u \cdot J_a(u_{j+1},v_{k}) = \dfrac{(1-\delta_u) \cdot \Delta J_{j-1,k}+ \Delta J_{j,k} + \delta_u \cdot \Delta J_{j+1,k}} {2} = \Delta_a J_{j,k} * w_u
\end{align*}
Where $w_u$ is a rectangular window function,
\begin{align*}
 w_u(x,y) = 
\begin{cases}
0.5 ,&  -1\leq x \leq1, y=0\\
0, & \text{otherwise.}
\end{cases}
\end{align*}
It is easy to show that similarly $E_2 = \Delta J_{j,k+1}* w_u $.
Thus, from linearity,
\begin{eqnarray*}
I^{B}_u(u,v) = I^{A}_{u}(u,v) * w_u
\end{eqnarray*}
And similarly for $v$, 
\begin{eqnarray*}
I^{B}_v(u,v) = I^{A}_{v}(u,v) * w_v
\end{eqnarray*}
Where $w_v$ is a rectangular window function,
\begin{eqnarray*}
w_v(x,y) &=& 
\begin{cases}
0.5 ,&  x=0,-1\leq y \leq1\\
0, & \text{otherwise.}
\end{cases}
\end{eqnarray*}

\section{Synthetic Color Transformation}
To simulate multimodality in the synthetic experiment, we apply a series of color effects to the RGB images in the ICL-NUIM dataset. 
The point clouds with the original colors are then aligned with the images whose colors have been modified.
The set of effects chosen was,
\begin{enumerate}
    \item Apply random color transformation of brightness, contrast, saturation, and hue using Pytorch ColorJitter with a random range of $[0, 0.4]^3$ and $[0,0.06]$, respectively.
    \item Apply gamma correction with a randomly selected gamma from $[0.5, 1]$ or $[1,2]$.
    \item Simulate different point spread functions and sensor properties by applying a Gaussian blur to the image using a random  $[0, 0.75]$.
\end{enumerate}
The next figure shows an example of an original image of the ICL-NUIM dataset (left) and the same image after applying such a random color transformation (right),

\begin{figure}[H]
    \centering
    \includegraphics[width=0.5\textwidth]{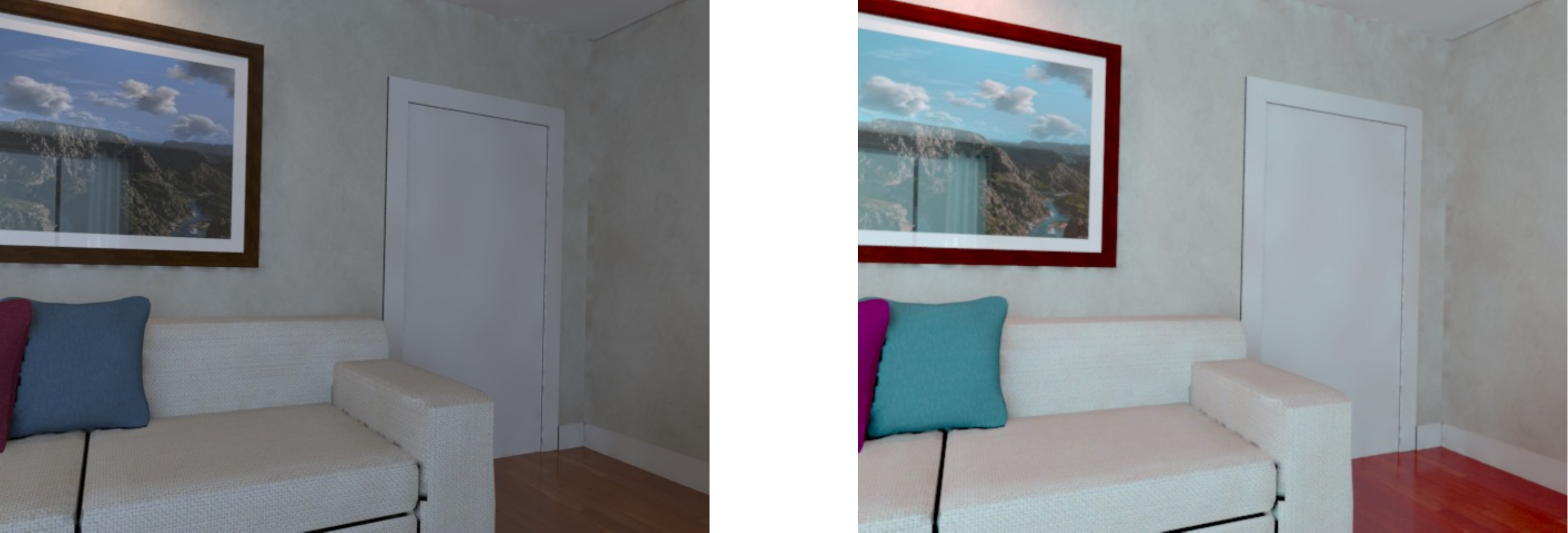}
    \label{fig:color_transform}
\end{figure}

\section{Synthetic Box Plot Results}
Section 
4.1 
presents cumulative normalized histograms of translation and rotation errors for synthetic data experiments.
Another informative way to present the results and effectiveness of our method is to present them in a boxplot.
The top row contains all comparisons from Section
4.1
, while the bottom row contains an enlarged version of the best performing methods. 
The proposed method with second-order polynomial color alignment
outperforms the other methods,
\begin{figure}[H]
    \centering
    \begin{minipage}[b]{0.6\textwidth}
    \includegraphics[width=\textwidth]{edges_comp/bp1.pdf}
    \label{fig:bp1}
  \end{minipage}    
   \centering
    \begin{minipage}[b]{0.6\textwidth}
    \vspace{-5mm}
    \includegraphics[width=1\textwidth]{edges_comp/bp2.pdf}
    \label{fig:bp2}
  \end{minipage}    
\end{figure}

\section{Robustness to Initialization}
To test the robustness of our method to initializations, the same experiment as in Section
4.1 
is performed with larger initialization values.
Instead of aligning each image in the ICL-NUIM dataset
with the subsequent image $I^{(i+1)}$ it is aligned with the non-consecutive image $I^{(i+3)}$.
As can be seen, this scenario maintains our favorable results,
\begin{figure}[H]
    \centering
    \includegraphics[width=0.7\textwidth]{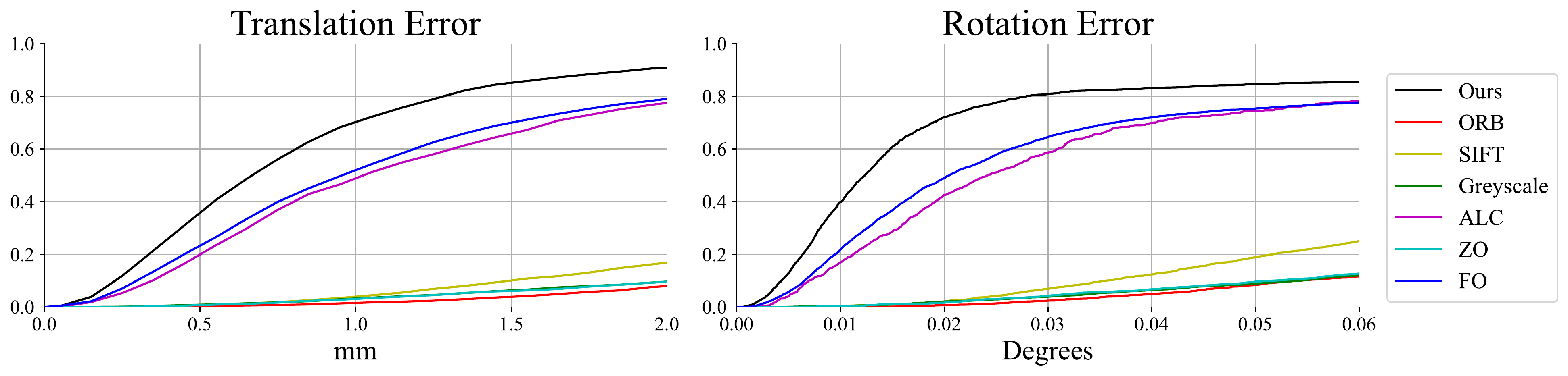}
    \label{fig:skip_results}
\end{figure}

\section{Third Degree Term}
As explained, in the study of Hong et al. for camera colorimetric characterization the third-order $RGB$ term is added as an additional dimension of the second-order polynomial kernel.
To test the advantage of adding this term, we perform the same experiment as in Section
4.1. 
The methods tested are the method proposed in the paper and the same method with the additional term.
As can be seen in the next figure, adding such a term has no significant effect on the results,
\begin{figure}[H]
    \centering
    \includegraphics[width=0.7\textwidth]{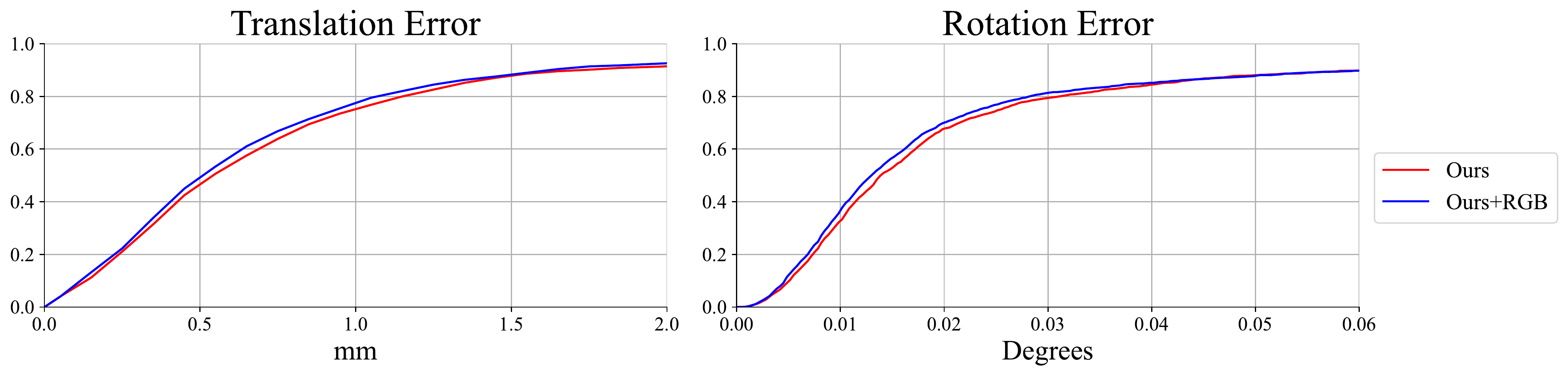}
    \label{fig:tdeg}
\end{figure}

\section{Real Data Results}
In the next figure, we present more comparisons between edges from camera images and edges extracted from rendered point cloud images, as in Figure 5,
\begin{figure}[H]
    \centering
    \includegraphics[width=\textwidth]{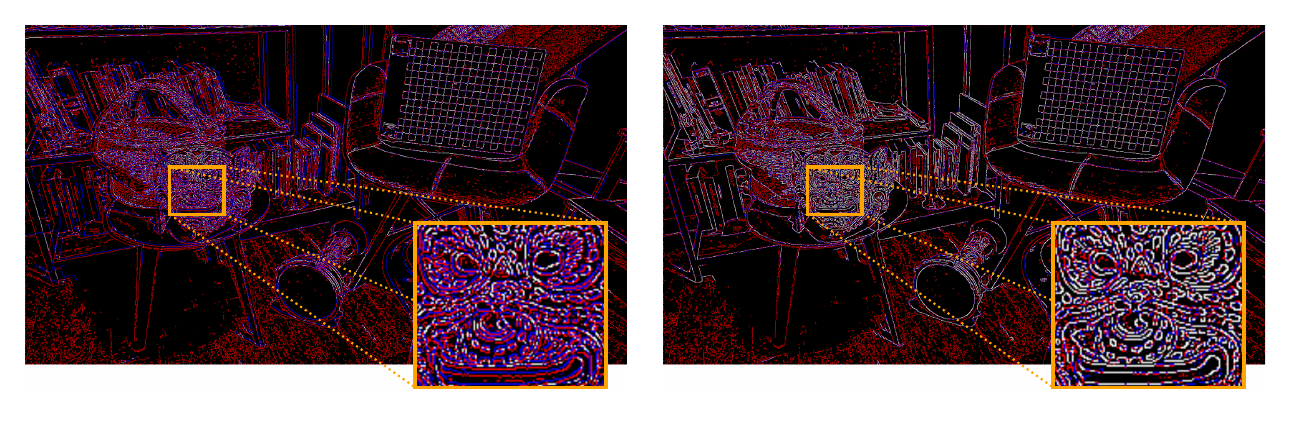}
    \label{fig:ec1}
\end{figure}
\begin{figure}[H]
    \centering
    \includegraphics[width=\textwidth]{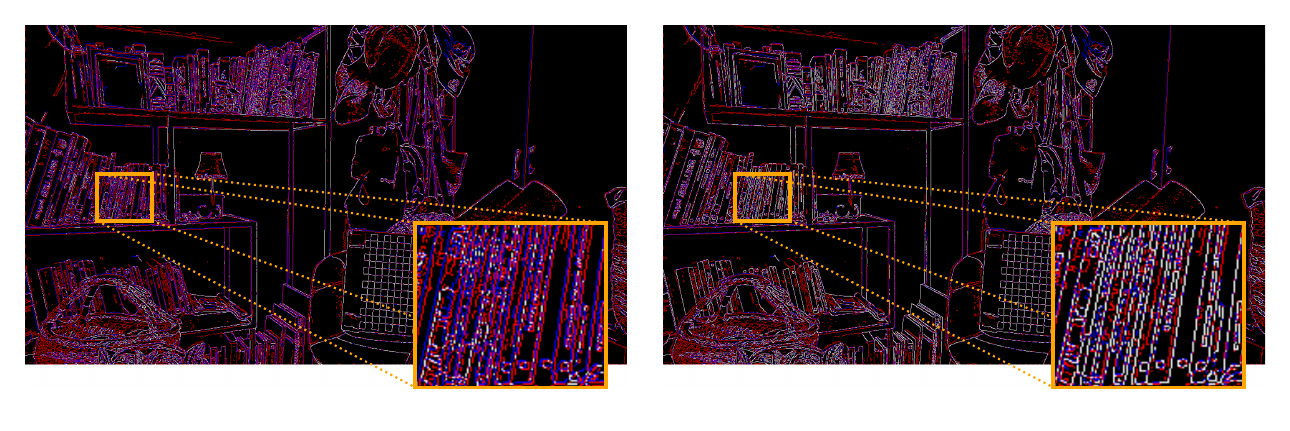}
    \label{fig:ec2}
\end{figure}
\begin{figure}[H]
    \centering
    \vspace{-16mm}
    \includegraphics[width=\textwidth]{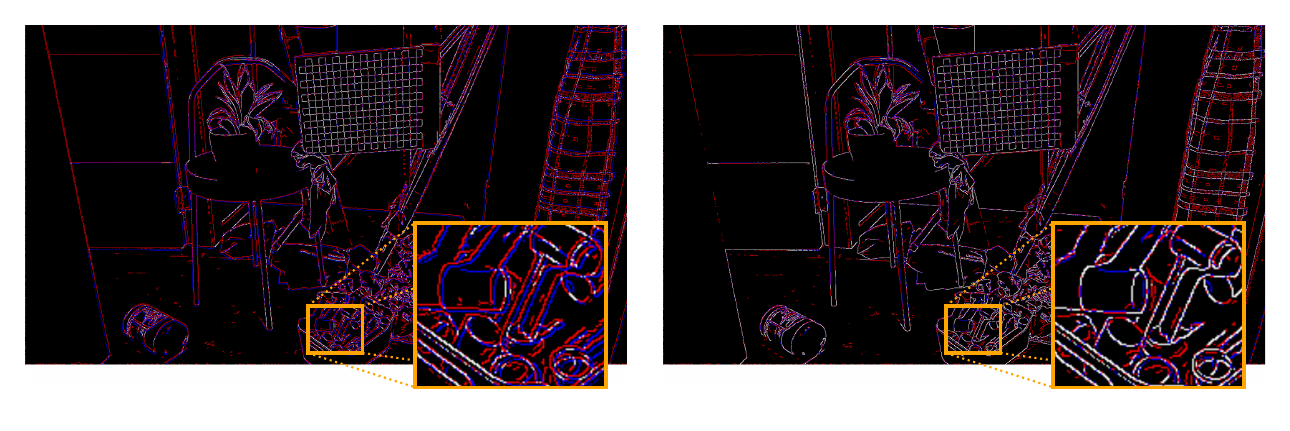}
    \label{fig:ec3}
\end{figure}
\begin{figure}[H]
    \centering
    \vspace{-16mm}
    \includegraphics[width=\textwidth]{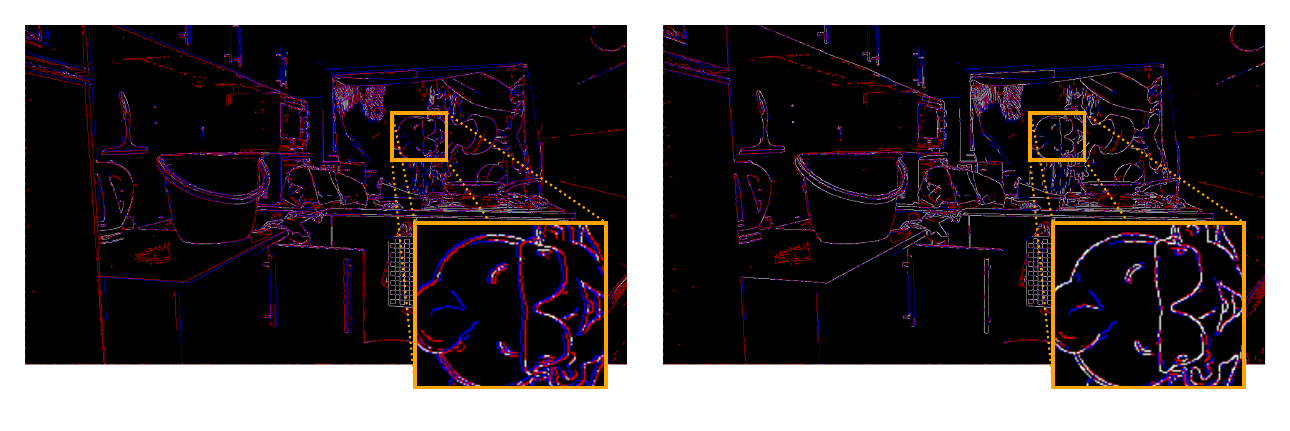}
    \label{fig:ec4}
\end{figure}
\begin{figure}[H]
    \centering
    \vspace{-16mm}
    \includegraphics[width=\textwidth]{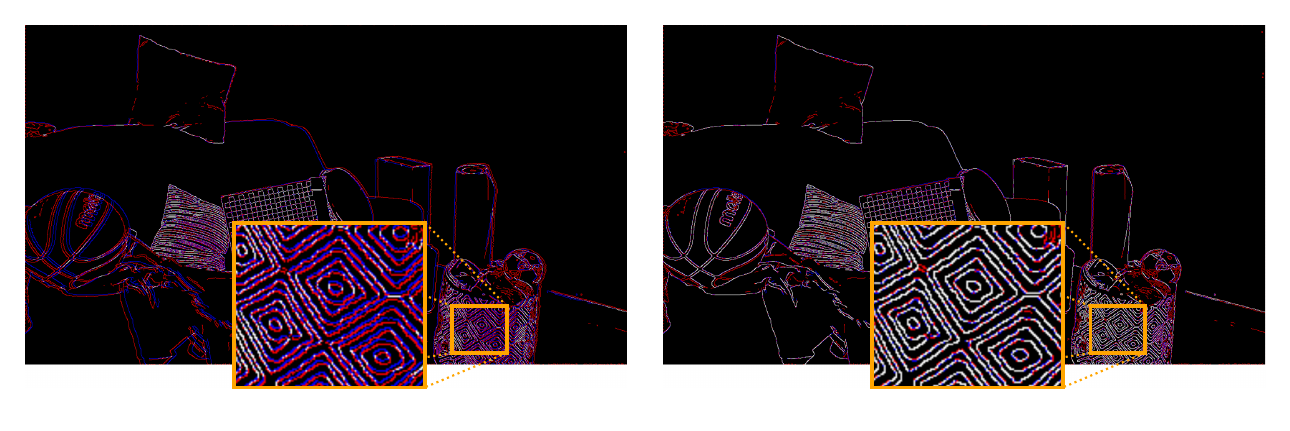}
    \label{fig:ec5}
\end{figure}
\begin{figure}[H]
  \centering
    \begin{minipage}[b]{\textwidth}
    \includegraphics[width=1\textwidth]{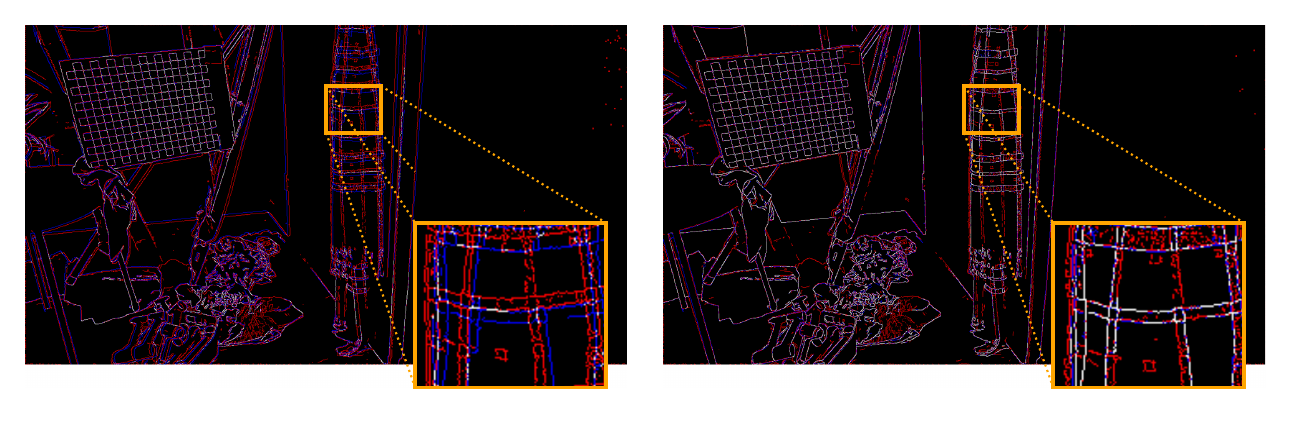}
    \label{fig:ec6}
  \end{minipage}    
\end{figure}
